# Predicting suicidal behavior among Indian adults using childhood trauma, mental health questionnaires and machine learning cascade ensembles

Akash K Rao[1][0000-0003-4025-1042], Gunjan Y Trivedi[2][0000-0003-4488-7945], Riri G Trivedi[2][0000-0001-5024-6103], Anshika Bajpai[1][0009-0007-4203-0581], Gajraj Singh Chauhan[1][0009-0000-7774-2887], Vishnu K Menon[1][0009-0007-9449-0934], Kathirvel Soundappan[3][0000-0002-4839-0138], Hemalatha Ramani[2][0009-0008-9513-8315], Neha Pandya[2][0009-0005-5981-3764], Varun Dutt[1][0000-0002-2151-8314]

[1] Applied Cognitive Science Laboratory, Indian Institute of Technology Mandi, Himachal Pradesh, India
[2] Society for Energy and Emotions, Wellness Space LLP, Ahmedabad, India
[3] Department of Community Medicine and School of Public Health, Post Graduate Institute of Medical Education and Research, Chandigarh, India
`akashrao.iitmandi@gmail.com`

**Abstract.** Among young adults, suicide is India's leading cause of death, accounting for an alarming national suicide rate of around 16%. In recent years, machine learning algorithms have emerged to predict suicidal behavior using various behavioral traits. But to date, the efficacy of machine learning algorithms in predicting suicidal behavior in the Indian context has not been explored in literature. In this study, different machine learning algorithms and ensembles were developed to predict suicide behavior based on childhood trauma, different mental health parameters, and other behavioral factors. The dataset was acquired from 391 individuals from a wellness center in India. Information regarding their childhood trauma, psychological wellness, and other mental health issues was acquired through standardized questionnaires. Results revealed that cascade ensemble learning methods using support vector machine, decision trees, and random forest were able to classify suicidal behavior with an accuracy of 95.04% using data from childhood trauma and mental health questionnaires. The study highlights the potential of using these machine learning ensembles to identify individuals with suicidal tendencies so that targeted interventions could be provided efficiently.

## 1 Introduction

Suicide is the most prevalent cause of death among the Indian population aged 15 to 49 [1]. India's proportion of worldwide suicide deaths rose among women from 25.3% in 1990 to 36.6% in 2016 and among men from 18.7% to 24.3% [1]. The research indicates that suicide is closely linked to mental health disorders (approximate-ly 90% of the individuals who commit suicide possess some mental health disorder, including psychiatric disorder, mood disorder, and substance use disorder, etc.) [3,7]. Suicide behavior (which includes suicidal thoughts and attempts) is defined as either having passive ideas about dying or an intention to kill oneself that is not accompanied by preparatory behavior. A recent systematic review found that mental disease increases the likelihood of suicide attempts tenfold [1]. To be effective at suicide prevention, it is vital to identify the risk factors. These risk factors include adverse childhood experiences (ACE), emotional abuse, a history of irrational decisions, inability to focus, and mental health issues [20-23]. Individuals who have undergone childhood trauma tend to have more mental and physical health issues, leading to a greater probability of suicide than individuals who have not undergone the trauma. Increased risk for suicide among those who experience child abuse or neglect has been reported in several meta-analysis results and replicated in numerous studies comprising different types of samples [1]. As a result, an investigation into the association between ACE exposure throughout childhood and adult mental health is crucial, with a possible connection to suicidal behavior [2].

Despite extraordinary initiatives to improve awareness and treatment, suicide rates have remained persistent and, in certain situations, have increased. Notably, most suicide victims meet with their doctors in the days and weeks following death [3], indicating missed risk detection despite treatment opportunities. Furthermore, recognized risk factors have a low sensitivity and therapeutic value in predicting suicide occurrence [3].

But in recent years, machine learning has provided researchers with an alternative to study and predict suicide. Machine learning algorithms strive to optimize prediction, whereas suicide theories seek to uncover causative factors [4,12]. To predict suicide behavior, machine learning algorithms frequently employ huge datasets- many variables with a large cohort of participants and exacerbated mathematical models. Researchers in [4] used machine learning classifiers like naïve bayes, support vector regression, elastic net penalized regression, and random forest to predict possible suicide attempts among US Army soldiers. Similarly, researchers in [5] employed natural language processing methods to identify psychiatric stressors leading to suicide from clinical notes to obtain an F-measure of 93%. Researchers in [6] employed a mixture of machine learning models and text mining methods to predict possible suicide attempts, using data acquired from the National Health Service (NHS) over three years. However, a detailed investigation on using machine learning models (and cascade ensemble learning methods) to predict suicide behavior is severely lacking and much needed in literature, especially in the Indian context. This work becomes even more important considering that India has the world's highest and most diverse population of early adults [1]. In addition, to the best of our knowledge, there is no literature on comparing ensemble cascade learning models and individual machine learning models in predicting suicide behavior. In this research work, we intend to address this gap in the literature by designing machine learning models for predicting suicide behavior among Indian adults by considering different psychological attributes acquired through questionnaires like the adverse childhood experiences (ACE), WHO-5 wellbeing index, insomnia severity index, major depression inventory, among others. In what follows, we give a detailed account of the dataset acquired at a wellness center in India. We then encapsulate the different machine learning models and the cascade ensemble learning models used in this work. We conclude by discussing our results and highlighting the implications of accurately predicting suicide behavior in the real world.

## 2 Methods

### 2.1 Dataset

The dataset was acquired from a wellness center based in Ahmedabad, India. The dataset acquired for the study conducted at the wellness center was based on IEC approval Ref No ECR/274/Inst/GJ/2013/RR-19, dated 27/10/2020. This dataset consisted of individuals who had voluntarily approached the wellness center for assistance with various mental health issues. All the individuals contacted the wellness center via social media or word of mouth and eventually requested a one-on-one consultation with the experts. Every participant agreed to pay for the consultation, signed a consent form, and was provided with videos about adverse childhood experiences and the consultation modus-operandi. They also filled out all necessary documents, such as signed consent, demographics, the adverse childhood experiences survey, and additional information.

The dataset consisted of 391 participants, where each row represents a participant, and the corresponding columns represented several attributes (in this case, the various questionnaires filled by the participants along with their answers to the questions asked in the interview in digitized form). These attributes consisted of a mixture of categorical questions (answered as yes/no), descriptive questions, and questions that were rated on a numeric scale. The various standardized questionnaires used for recording adverse childhood experiences and other mental health issues filled by the participants are as shown in Table 1.

**Table 1.** Different standardized questionnaires used in the dataset.

| Questionnaire | What does it measure | Type of measure |
|---|---|---|
| Adverse Childhood Experiences (ACE) questionnaire [2] | This questionnaire consists of 16 questions. It covered adverse childhood experiences from the age of 3 to 18. Core categories of abuse include abuse, neglect, household experiences, peer violence, and bullying. | Binary (Rated as Yes/No). "Yes" is encoded as 1, and a "No" is encoded as a 0, and the cumulative score is considered. |
| WHO-5 Well-being index [8] | This questionnaire consists of 5 questions. It measures subjective psychological wellbeing based on optimism, vitality, and general interest. All the questions are asked from the perspective of the participant's wellbeing over the past 14 days. | It was measured over a 6-point Likert scale, ranging from "At no time" (encoded as 0) to "All the time" (encoded as 5). The total score is |

| | | eventually multiplied by a factor of 4 to obtain a percentage score. |
|---|---|---|
| Major Depression Inventory (MDI) [9] | This questionnaire consists of 12 questions. It measures the level of depressive symptoms among patients with questions on self-confidence, guilt, sleep, restlessness, appetite, and energy. | It was measured over a 6-point Likert scale, ranging from "At no time" (encoded as 0) to "All the time" (encoded as 5). The total score is taken into consideration. |
| General Anxiety Disorder (GAD) – 7 [10] | This questionnaire consists of 7 questions. It measures anxiety symptoms with questions on the extent of relaxation, state of worrying, and restlessness, among others. All the questions are asked from the perspective of the participant's wellbeing over the past 14 days. | It was measured over a 4-point scale, ranging from "Not at all sure" (encoded as 0) to "Nearly every day" (encoded as 3). The total score was taken into consideration. |
| Insomnia Severity Index (ISI) [11] | This questionnaire consists of 7 questions. It is a self-reported questionnaire to gauge how severe insomnia is in adults. The questionnaire consists of measures like daytime functioning, sleep pleasure, and perceived severity of sleep issues. | It was measured over a 5-point scale, ranging from "None" (encoded as 0) to "Very severe" (encoded as 4). The total score is taken into consideration. |

As mentioned above, in addition to the standardized questionnaires, the interviewer also asked questions based on the participant's psychological turmoil. These questions are encapsulated in Table 2.

**Table 2.** Questions based on the individual's psychological turmoil.

| Questions | Explanation and type of measure |
|---|---|
| Have you attempted any self-harm – cutting yourself, banging head, etc.? | Self-harm can be described as a purposeful, direct injury or change to body tissue that is not done to commit suicide. It is encoded in a binary format (Yes/No). |
| Any irrational decisions that may have caused long-term impact on your lives? | Irrational decisions encompass those decisions that are made without rational reasoning. It is encoded in a binary format (Yes/No). |
| Are you unable to focus/concentrate? | This question examines any chronic difficulties with the individual's capacity to focus or concentrate. It is encoded in a binary format (Yes/No). |
| Have you had any | This question inquires whether the individual has any ongoing |
| relationship issues with your parents/siblings? | relationship problems with their parents, siblings, or spouse. It is encoded in a binary format (Yes/No). |
| Are you physically violent towards others? | This question inquires whether the individual has a history of being physically violent towards others. It is encoded in a binary format (Yes/No). |
| Have you grown up in a violent family environment? | This question inquires whether the individuals have grown up in a violent family environment. It is encoded in a binary format (Yes/No). |

Suicidal behavior was taken as the study's output variable. This question was coded in a binary form, where "Yes" indicated either a history of suicide behavior or a history of suicide attempt, and "No" stated neither. The difference between the aforementioned suicidal behaviors was inspired from [20], who reasoned that nonfatal suicidal thoughts and behaviors can be more precisely classified into three distinct groups: suicide ideation, which encompasses thoughts of engaging in behavior aimed at ending one's life; suicide plan, which corresponds to the establishment of a particular procedure through which one intends to die; and suicide attempt, which alludes to engagement in potentially harmful behavior with at least some intent to die [20]. A preprocessing pipeline was developed using custom codes, which included a power transformer to transform skewed columns and standardize the remaining columns.

## 2.2 Machine learning algorithms

**Support Vector Classifier (SVC).** SVM is a supervised machine learning algorithm that finds a hyperplane that creates a boundary between data categories. This hyperplane is a line in two dimensions [6].

**Multi-layer perceptron (MLP).** An MLP classifier is a feedforward neural network comprising several neuron layers. Each layer is connected to the next layer by edges labeled with weights [3].

**Decision trees (DT).** A DT is a tree-like model in which the interior nodes symbolize feature tests, the branches indicate test results, and the leaf nodes represent final choices or predictions [3]. The algorithm may deconstruct a vast and complex decision-making process into smaller, simpler ones, making the results easier to grasp and analyze [3].

**Random Forest (RF).** The RF creates numerous decision trees based on a random subset of the input features and a random portion of the training data. The definitive forecast is then created by averaging all the tree predictions [3,15]. As a result, the individual trees are less likely to match the noise in the data and more likely to identify the underlying patterns, lowering the danger of overfitting [3,15].

**Cascading ensembles.** In cascading ensembles, individual classifiers are trained using 10-fold cross-validation to classify suicide behavior in CE models [14]. The best of the 10-folds was eventually used for classification. Following that, the learned classifiers' predictions are utilized as attributes alongside other attributes in the data and a second model is trained to classify suicide behavior [14]. Thus, in this method, the subsequent model uses the earlier models' expertise to increase classification accuracy. In SVC-DT Ensemble, SVC traits were combined with additional features to form a DT model. SVC and DT features were combined to create an RF model in SVC-DT-RF Ensemble.

The different hyperparameters and the corresponding values used in different machine learning algorithms are as shown in Table 3.

**Table 3.** Different hyperparameters and their corresponding range of values used in different machine learning algorithms.

| Machine learning model | Hyperparameters and their values |
|---|---|
| Support Vector Classifier (SVC) | Kernel Function - Linear, poly, radial basis function, sigmoid<br>Gamma – Scale, auto<br>C – 0.5, 1, 2<br>Degree of the polynomial kernel – 2, 3, 4<br>Probability - True |
| Multi-layer perceptron | Number of neurons in the hidden layer – 32, 64, 128<br>Number of hidden layers – 2, 3, 4<br>Activation function – Relu, Tanh<br>Solver – sigmoid, adam<br>Alpha – $1e^{-4}$, $3e^{-4}$, 0.05<br>Learning rate – Constant, adaptive |
| Decision trees and Random Forest | Tree criteria – Gini, Entropy<br>Maximum tree depth – 2, 3, 5, 10, 50<br>Maximum feature function – Square root, $\log_2$<br>Minimum samples at leaf node – 1, 5, 8, 10<br>Minimum samples for split – 2, 3, 50, 100 |

**Feature selection.** A sequential feature selector (SFS) was used to identify the most relevant features in the dataset. Based on the accuracy metric, the SFS analyzed the performance of multiple models using different subsets of information. The algorithm assessed the model's performance with each iteration added or removed feature and chose the next best feature based on the outcomes of the previous iteration [3].

**Cross-validation.** A data set is divided into numerous folds or subsets in cross-validation (the number of folds is "K" in K-fold Cross Validation), one of which is utilized as a validation set. The leftover folds, on the other hand, are utilized to train the model. This division is done numerous times, with the validation set changing each time. The results of each validation phase are averaged to produce a more accurate approximation of the model's performance. To train and test the machine learning models, this work used 10-fold cross-validation [13].

**Error matrices.** We intended to classify the existence of suicide behavior (including thoughts/attempt) based on all the attributes mentioned above, making this a two-class classification problem. The error matrices included:

*Accuracy.* The accuracy is defined as:

$$Accuracy = \frac{TP}{TP+FN+TN+FP} \quad (1)$$

Where TP is the number of true positives (suicide behavior present as "Yes"), TN is the number of true negatives (suicide behavior absent as "No"), FP is the number of false positives (suicide behavior absent as "Yes"), and FN is the number of false negatives (suicide behavior present as "No").

*Sensitivity.* Sensitivity is calculated as:

$$Sensitivity = \frac{TP}{TP + FN}$$

*Specificity:* The specificity was calculated as:

$$Specificity = \frac{TN}{TN + FP}$$

Accuracy was used as a measure for optimizing models in this research. Following feature selection, the models were optimized using a grid search procedure that selected the best set of hyperparameters to maximize the cross-validation test accuracy.

## 3   Results

Table 4 shows the different machine learning models employed to predict suicide behavior, the best set of hyperparameters obtained for each algorithm, and the selected feature in each machine learning algorithm.

**Table 4.** Different machine learning models employed along the features selected.

| Category of the parameter | Parameter description | Support Vector Classifier (SVC) | Multi Layer Perceptron (MLP) | Decision Trees (DT) | Random Forest (RF) | SVC-DT cascade ensemble | SVC-DT-RF cascade ensemble |
|---|---|---|---|---|---|---|---|
| Childhood Trauma (Adverse Childhood Experiences-ACE) | Emotional Abuse | ✓ | ✓ | ✓ |  | ✓ | ✓ |
|  | Emotional Neglect | ✓ | ✓ | ✓ |  | ✓ | ✓ |
|  | Witness to domestic volence | ✓ | ✓ | ✓ |  | ✓ | ✓ |
|  | Parent's Mental Health |  | ✓ |  |  |  | ✓ |
|  | Peer Rejection | ✓ | ✓ |  |  | ✓ | ✓ |
|  | Parents Fighting | ✓ | ✓ | ✓ |  | ✓ | ✓ |
|  | **Cumulative ACE score** | ✓ | ✓ | ✓ | ✓ | ✓ | ✓ |
| Mental health parameters | WellBeing ( WHO 5) |  |  | ✓ |  | ✓ | ✓ |
|  | Anxiety (GAD 7) | ✓ | ✓ |  | ✓ | ✓ | ✓ |
|  | Depression (MDI) | ✓ | ✓ |  |  | ✓ | ✓ |
|  | Insomnia (ISI) | ✓ |  | ✓ | ✓ | ✓ | ✓ |
| Behavioral factors | History of Self Harm |  |  |  |  | ✓ | ✓ |
|  | Irrational Decisions | ✓ |  |  |  | ✓ | ✓ |
|  | Ability to Focus |  |  |  | ✓ | ✓ | ✓ |
|  | History of Violent Decisions |  |  | ✓ | ✓ |  |  |
|  | Relationship Issues |  | ✓ |  |  |  |  |

**Table 5.** Different machine learning models employed along with the best set of hyperparameters obtained.

| Machine learning model | Optimal hyperparameters |
|---|---|
| Support vector classifier (SVC) | C = 2, degree = 2, gamma = scale, kernel = radial basis function, probability = true |
| Multi-layer perceptron (MLP) | Activation = relu, alpha = 0.0003, hidden layer sizes = (64, 64, 64), learning rate = adaptive, solver = rgd |
| Decision Trees (DT) | Criterion = gini, maximum depth = 50, samples split = 2, minimum samples in leaf node = 5 |
| Random Forest (RF) | Criterion = entropy, maximum depth = 2, samples split = 50, minimum samples in leaf node = 8 |
| SVC-DT cascade ensemble | Same as SVC parameters |
| SVC-DT-RF ensemble | Same as SVC and DT parameters |

Table 6 shows the cross-validation accuracies, sensitivities, and specificities, for different machine learning models during training and testing. As can be observed, the SVC-DT-RF cascade ensemble learning algorithms significantly outperformed the other machine learning algorithms. The cascade ensemble learning models did not overfit (as can be observed, the difference between training and testing accuracies was meager). Based upon the attributes obtained in cascade ensemble learning models, it appeared that an individual's suicide behavior was largely dependent on the childhood trauma (acquired through the ACE-16 questionnaire), mental health parameters

(WHO-5, GAD-7, ISI, and MDI) and behavioral factors (history of self-harm, irrational decisions, and ability to focus).

**Table 6.** Different machine learning models and their best cross-validation accuracies obtained in training and testing.

| Machine learning model | Cross-validation training accuracy | Cross-validation test accuracy | Cross-validation training sensitivity | Cross-validation test sensitivity | Cross-validation training specificity | Cross-validation test specificity |
|---|---|---|---|---|---|---|
| Support vector classifier | 73.57 | 67.51 | 54.67 | 50.88 | 87.66 | 80.49 |
| Multi-layer perceptron | 67.03 | 68.31 | 51.77 | 51.03 | 78.52 | 81.78 |
| Decision trees | 70.56 | 64.67 | 60.57 | 54.74 | 78.12 | 71.23 |
| Random Forest | 78.65 | 73.45 | 53.45 | 39.81 | 87.66 | 77.56 |
| SVC-DT cascade ensemble | 87.66 | 81.55 | 79.55 | 73.45 | 89.23 | 82.33 |
| SVC-DT-RF cascade ensemble | **96.76** | **95.04** | 87.88 | 86.75 | 93.22 | 91.29 |

## 4　Discussion and conclusion

This research aimed to efficiently predict suicide behavior (including thoughts and attempts) using different attributes (like childhood trauma/adverse childhood experiences, mental health parameters, internalization issues, behavioral factors, and externalization issues) and machine learning/cascade ensemble learning algorithms. The results revealed that the cascade ensemble learning algorithms comfortably outperformed their machine learning counterparts in accurately predicting suicide behavior. These results were consistent with the results obtained in [14], who reasoned that since cascade ensemble learning models effectively reduce bias and variance errors in the data, it usually leads to a higher accuracy while testing. The results also revealed that cascade ensemble learning approaches were less sensitive to the outliers and noise in the dataset (usually in datasets involving subjective ratings recorded from human participants) and hence were more robust. As suggested in [14], the cascade ensemble learning algorithms can alleviate possible overfitting due to their very nature, thereby reducing the risk of any single model (in this case, SVC, DT, and RF) overfitting to the training set's idiosyncrasies. This research also integrated three categories (childhood trauma, mental health parameters, and behavioral factors) and several parameters including cumulative trauma for predicting suicidal behavior and obtained a better classification accuracy than some of the previous research studies [16-19].

The research work also found out the individual's suicide behavior was dependent on factors like depression, anxiety, insomnia, the total score in the ACE questionnaire (encapsulating the presence of several trauma elements), history of self-harm, and the WHO-5 Wellbeing index among others. Specifically, a combination of ACE questions based on emotional abuse, peer isolation, witness to domestic violence contributed significantly to model classification. This result has important clinical implications for clinical practice and formulating health policies in the Indian context. Identifying individuals with high depression levels, anxiety, insomnia, and ACE scores might go a long way in providing early customized interventions and support to individuals. However, the research work does have some limitations. Even though the dataset was acquired over 6 months, the sample size used in the work is rather limited (N = 391). It might not fully represent the complexities and heterogeneity of the general population (especially a population as diverse as India). Since the study was cross-sectional, it was impossible to determine the causality or direction of the relationship between suicidal behavior and psychological attributes.

In the future, we intend to expand our horizons and acquire datasets from multiple wellness centers and hospitals in India to test the validity and generalizability of our models. In addition, we also intend to evaluate the effects of substance abuse, occupation, comorbidities, and other socio-economic characteristics on suicide behavior. This might lead to the early identification of suicidal behavior so that appropriate assistance and interventions can be provided.